%% file: main.tex
\documentclass{article} 
\usepackage{iclr2027_conference,times}

\input{math_commands.tex}

\usepackage{amsmath}
\usepackage[ruled,linesnumbered]{algorithm2e}
\usepackage{afterpage}
\usepackage{graphicx}
\usepackage{multirow}
\usepackage[utf8]{inputenc} 
\usepackage[T1]{fontenc}    
\usepackage{hyperref}       
\usepackage{url}            
\usepackage{booktabs}       
\usepackage{amsfonts}       
\usepackage{nicefrac}       
\usepackage{microtype}      
\usepackage{xcolor}         

\title{\textsc{TAME}: Topology-Aware Text-Driven Motion Editing across Heterogeneous Humanoid Skeletons}

\author{\textbf{Qichen Zheng$^{1}$, Siyuan Yang$^{2}$, Chong Wang$^{1}$, Jun Liu$^{3}$,} \\
\textbf{Shijian Lu$^{1}$, Alex Kot$^{1}$, Kwok-Yan Lam$^{1}$} \\[2pt]
{\normalfont $^{1}$Nanyang Technological University} \\
{\normalfont $^{2}$KTH Royal Institute of Technology \qquad $^{3}$Lancaster University} \\
{\normalfont\texttt{qichen001@e.ntu.edu.sg}}
}

\iclrfinalcopy 
\begin{document}

\maketitle

\afterpage{%
\begin{figure}[t]
  \centering
  \includegraphics[width=\linewidth]{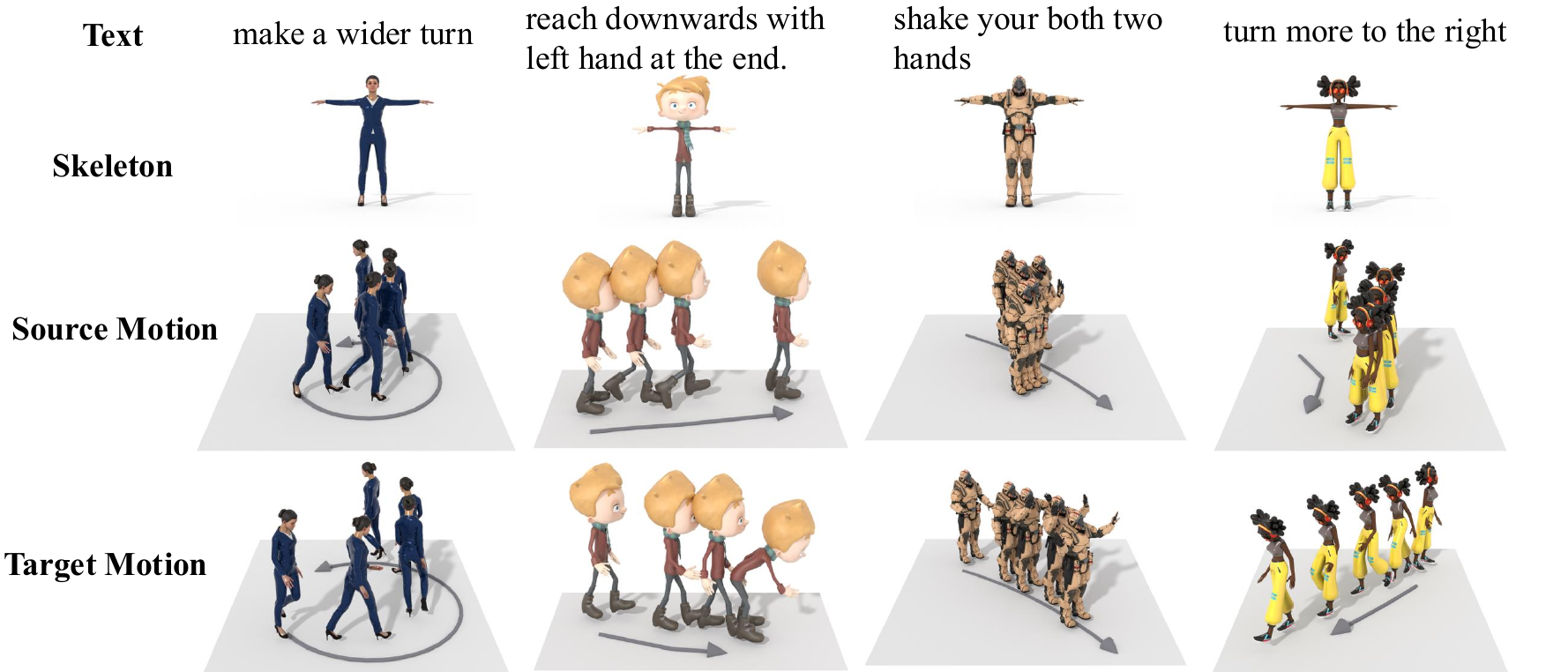}
  \caption{Text-driven motion editing across heterogeneous humanoid
  skeletons. \textsc{TAME} generates edited motions from a text
  instruction, a source motion, and the character's skeleton topology.}
  \label{fig:task}
\end{figure}
}

\afterpage{\afterpage{\afterpage{\afterpage{%
\begin{figure}[t]
  \centering
  \includegraphics[width=\linewidth]{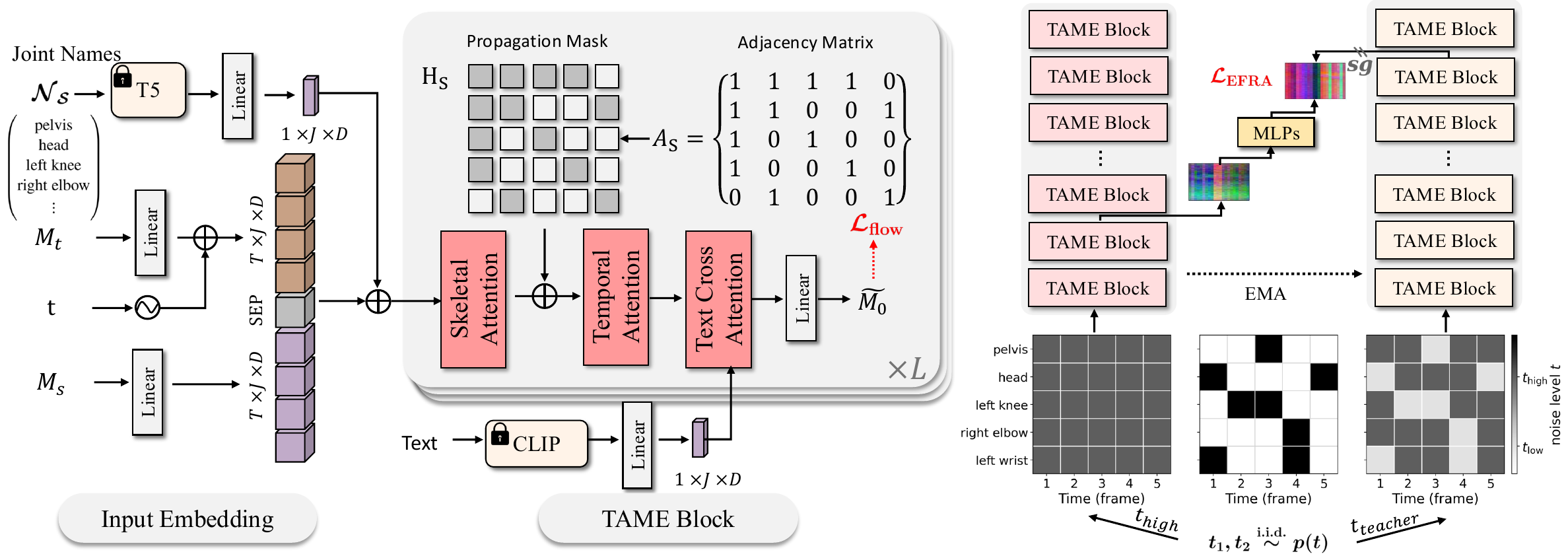}
  \caption{Overview of \textsc{TAME}. \textbf{Left:} the
  topology-aware motion editor backbone embeds the source motion
  $M_s$ and noised target motion $\tilde{M}_t$ as joint-time tokens
  and applies $L$ layers of topology-constrained skeletal
  propagation (\textsc{TCSP}), temporal self-attention, and text
  cross-attention to predict the velocity field $\hat{v}_\theta$.
  \textbf{Right:} the Edit-Focused Representation Alignment
  (\textsc{EFRA}) branch, used only in training. A DTW edit map $E$
  marks the edit-relevant tokens and drives edit-aware dual-timestep
  scheduling, which makes the EMA teacher cleaner than the student
  on those tokens. Student and teacher features are then aligned
  only on those tokens.}
  \label{fig:method}
\end{figure}
}}}}

\begin{abstract}
Text-driven motion editing modifies an existing motion sequence
according to a text instruction while preserving the
content of the source motion. Existing methods are typically built
for a single, fixed skeletal topology, which limits their use in
animation pipelines where characters differ in joint count and
skeletal hierarchy. We present \textsc{Topology-Aware Motion Editor}
(\textsc{TAME}), a flow-matching transformer that edits motions
on humanoid skeletons of varying topology. \textsc{TAME}
represents motion as per-joint, per-frame tokens and models
interactions among joints, across frames, and with the text
instruction through skeletal, temporal, and text cross-attention
layers. To make the skeletal attention follow each character's
hierarchy, \textsc{TAME} replaces full joint attention with
Topology-Constrained Skeletal Propagation (\textsc{TCSP}), which
restricts attention to one-hop kinematic neighbors in the
skeleton's adjacency matrix. 
We further introduce Edit-Focused Representation Alignment (\textsc{EFRA}), a self-distilled representation alignment strategy that aligns student features with cleaner EMA-teacher features exclusively on edit-relevant joint-time tokens, making edits faithful to the instruction.
To make this setting trainable and comparable, we construct
\textsc{TopoMotionFix}, a multi-topology extension of MotionFix with
seen- and unseen-topology evaluation protocols. \textsc{TAME}
outperforms previous methods in edit alignment and
source preservation on MotionFix and reliably edits motions on
unseen skeletons in \textsc{TopoMotionFix}.
\end{abstract}

\section{Introduction}
\label{sec:introduction}

Character animation~\citep{tevethuman,zhang2024motiondiffuse,guo2024momask} is rarely
completed in a single generation pass: an initial motion may not match the creator's intent
due to the ambiguity of high-level language instructions. Motion
editing~\citep{athanasiou2024motionfix,delmas2023posefix} therefore plays a crucial role in
animation workflows: it refines an existing motion according to a
text instruction, such as raising an arm higher, while keeping
unedited motion unchanged.


In animation production, a motion editor is expected to accommodate
the varying skeletal topologies of different characters and to edit a
motion faithfully according to the text instruction. Despite recent
progress in text-driven motion
editing~\citep{li2025simmotionedit,motionclr,yang2025partmotionedit},
existing methods fall short in both respects. First, most existing
methods overlook the broader need of diverse character animation:
editing motions across different skeletal topologies. Instead, they
are designed for a single fixed
topology~\citep{athanasiou2024motionfix}. However, in real-world
production
environments~\citep{gat2025anytop,lee2023same,yang2024omnimotiongpt},
human motion data commonly involves varying skeletal configurations
beyond the canonical topology assumed by these methods. This topology
mismatch limits the practical applicability of current motion editing
methods. 
 Second, existing methods struggle to realize the requested edit, so the edited motion is often misaligned with the instruction.
For instance, when the
required change is subtle, as in ``stand without stretching
leg,'' current methods tend to ignore it, whereas
when the instruction is complex or abstract, as in ``act as if violin
is being played,'' the generated motion deviates substantially from
the text semantics. Applying the requested edit while preserving the
unedited motion remains an unsolved
challenge.

To edit motion across different skeletal topologies while keeping
each edit faithful to the instruction, we propose \textsc{TAME} (Topology-Aware
Motion Editor), a flow-matching~\citep{lipmanflow} transformer for
text-driven motion editing.
Existing motion editing
methods~\citep{athanasiou2024motionfix,li2025simmotionedit,delmas2023posefix,motionclr} typically
represent each pose as a single vector and focus mainly on temporal
relationships while overlooking spatial interactions
among skeletal joints. To address this
limitation, \textsc{TAME} embeds motion as per-joint tokens at each
frame, forming a skeleto-temporally structured representation that
keeps the per-joint feature dimension fixed as the number of joints
varies across skeletons. The network applies attention along
the skeletal and temporal axes to model joint interactions and motion
dynamics, while text cross-attention captures fine-grained
interactions between the edit prompt and skeleto-temporal motion
units. To handle motions on skeletons of varying topology, we
introduce \textsc{Topology-Constrained Skeletal Propagation}
(\textsc{TCSP}) for skeletal interaction: given the input topology,
\textsc{TCSP} permits one-step communication only between a joint and
its physical neighbors, so that topology becomes a structural
communication rule rather than a learnable bias in full
attention. Long-range coordination remains possible but must emerge
through multi-layer propagation along the character's kinematic chain.

To make edits faithful to the instruction, we propose Edit-Focused
Representation Alignment (\textsc{EFRA}), which imposes
representation guidance on the edited parts of a motion
(Fig.~\ref{fig:method}). Our key insight is that motion editing is
an asymmetric learning problem: the source motion provides a direct
reference for the unedited parts, whereas the edited parts can only
be inferred from its interaction with the text instruction. The
edited parts therefore call for stronger guidance. Representation
alignment lends itself to this purpose: recent work has shown that a
diffusion transformer generates better when its intermediate
features are aligned to a stronger representation, and that this
representation can come not only from an external
encoder~\citep{yurepresentation} but also from the network itself,
whose features grow more informative as the noise decreases and the
depth increases~\citep{jiang2025no}. Accordingly, \textsc{EFRA}
realizes this guidance by self-distillation and applies the
alignment to the edited parts. It aligns an earlier-layer student
feature under higher noise to a later-layer teacher feature under
lower noise. The alignment is restricted to the edit-relevant
joint-time tokens, identified during training by a DTW edit map
derived from each source--target pair. The teacher is an EMA copy
of the student and is the network deployed at inference, so
\textsc{EFRA} adds no inference cost.


To train and evaluate \textsc{TAME} on heterogeneous humanoid
skeletons, we construct \textsc{TopoMotionFix}, a multi-topology
extension of MotionFix~\citep{athanasiou2024motionfix} that
retargets its source--target--instruction triplets onto a pool of
humanoid rigs with different joint counts and hierarchies, and
provides both seen- and unseen-topology evaluation protocols.

In summary, our contributions are fourfold. First, we propose
\textsc{TAME}, to the best of our knowledge the first text-driven
motion editor for heterogeneous humanoid skeletons, whose
topology-constrained skeletal propagation (\textsc{TCSP}) turns the
input topology into a structural communication rule of the network.
Second, we introduce \textsc{EFRA}, a self-distillation strategy
that strengthens guidance on the edited parts of a motion and
thereby makes edits more faithful to the instruction. Third, we
construct
\textsc{TopoMotionFix}, a multi-topology benchmark built on
MotionFix with seen- and unseen-topology evaluation protocols.
Fourth, \textsc{TAME} outperforms prior methods on MotionFix in
both edit alignment and source preservation and generalizes robustly
to unseen skeleton topologies in \textsc{TopoMotionFix}, with
ablations validating the contributions of \textsc{TCSP} and
\textsc{EFRA}.

\section{Related Work}
\label{sec:related_work}

\subsection{Text-Driven Motion Editing}
\label{subsec:rw_motion_editing}

Early 3D motion editing relied on handcrafted spatial-temporal
constraints to manipulate
trajectories~\citep{witkin1988spacetime,gleicher1997motion}, where
each edit demands manually specified rules and does not generalize
to free-form text instructions. Motion editing methods including
DNO~\citep{karunratanakul2024optimizing},
MoMo~\citep{raab2024monkey}, MotionCLR~\citep{motionclr}, and
SALAD~\citep{hong2025salad} manipulate the attention or initial
noise of pretrained motion diffusion models without further
training; in the absence of paired editing supervision, however,
these methods struggle to deliver fine-grained semantic edits while
preserving the source motion. LLM-driven editors such as Iterative
Motion Editing~\citep{goel2024iterative} and
CoMo~\citep{huang2024controllable} translate text into symbolic
operations applied to a frozen motion model; however, they rely on
a textual description of the source motion as input rather than the
motion data itself. MotionFix~\citep{athanasiou2024motionfix}
introduces the canonical triplet-based supervised setting (source
motion, target motion, instruction) for text-driven 3D motion
editing. Building on it, SimMotionEdit~\citep{li2025simmotionedit}
and PartMotionEdit~\citep{yang2025partmotionedit} add motion
similarity prediction as an auxiliary task at the frame and part
level, respectively, while MotionReFit~\citep{jiang2025dynamic}
blends source and target motions through the MotionCutMix
augmentation; more recently, OmniME~\citep{Shi_2026_CVPR} adds
negative supervision through a text--motion triplet loss, and \citet{Han_2026_CVPR} fuse frame-wise and joint-wise motion
features and predict joint-wise motion differences as an auxiliary
task. These methods, however, are designed for a single fixed
topology and struggle to realize the edit instruction faithfully.
\textsc{TAME} addresses both limitations with a topology-aware
editing architecture that generalizes across heterogeneous
humanoid skeletons and a self-distilled representation alignment
that provides more informative guidance on the edited regions.

\subsection{Skeleton-Agnostic and Arbitrary-Topology Motion Modeling}
\label{subsec:rw_topology}

Existing methods modeling motion across diverse skeletal
structures~\citep{lee2023same,yang2024omnimotiongpt,gat2025anytop,liu2025text,khani2025unimogen,mo2025pumps,wang2026x,chen2025topology,xu2026necromancer}
focus on generation, representation, or retargeting rather than
editing. Early work such as SAME~\citep{lee2023same} learns a
skeleton-agnostic motion embedding, and
OmniMotionGPT~\citep{yang2024omnimotiongpt} extends generation to
animal skeletons. Recent generation works further diversify
backbones and data scale, including AnyTop~\citep{gat2025anytop}
with a Graphormer-style attention
bias~\citep{ying2021transformers}, UniMoGen~\citep{khani2025unimogen},
OmniSkel~\citep{liu2025text}, PUMPS~\citep{mo2025pumps},
X-MoGen~\citep{wang2026x}, OmniZoo~\citep{chen2025topology}, and
NECromancer~\citep{xu2026necromancer}. A separate retargeting thread
pioneered by skeleton-aware
networks~\citep{aberman2020skeleton} and
Motion2Motion~\citep{chen2025motion2motion} transfers a given motion
across skeletons rather than editing it. Architecturally, prior methods such as AnyTop~\citep{gat2025anytop}
inject topology through attention biases that allow long-range
joint communication within a single layer. \textsc{TAME} instead
enforces a hard one-hop adjacency mask (\textsc{TCSP}), so that
distant joints interact only via multi-layer propagation along the
kinematic chain.

\subsection{Representation Alignment in Diffusion Models}
\label{subsec:rw_repr_align}

Representation alignment has emerged as an effective auxiliary
regulariser for diffusion transformer training.
REPA~\citep{yurepresentation} aligns the noisy hidden states of the
denoiser with clean features from a pretrained visual encoder.
Subsequent work traces its gain to spatial structure rather than global
semantics~\citep{singh2025matters}, extends alignment to pixel-space
transformers~\citep{shin2026representation,wang2026persistence}, and
builds the encoder into the latent space itself~\citep{zheng2026diffusion}.
SRA~\citep{jiang2025no} removes the external dependence by
self-distilling across noise levels, matching an earlier student
feature under higher noise to a later EMA-teacher feature under
lower noise. To the best of our knowledge, \textsc{EFRA} is the first to apply
representation alignment to text-driven motion editing, through a
token-wise dual-timestep schedule driven by a DTW-derived edit
mask.

\section{Method}
\label{sec:method}

\subsection{Overview}
\label{subsec:overview}

Given a source motion $M_s \in \mathbb{R}^{T \times J \times D}$
defined on a skeleton structure $\mathcal{S}$ and a text editing
instruction $c$, our goal is to generate an edited motion
$M_t \in \mathbb{R}^{T \times J \times D}$ that semantically aligns
with the instruction while preserving source motion content. The
skeleton structure is represented as
$\mathcal{S}=\{\mathcal{N}_{\mathcal{S}}, A_{\mathcal{S}}\}$, where
$\mathcal{N}_{\mathcal{S}}=\{n_j\}_{j=1}^{J}$ is the set of joint
names and $A_{\mathcal{S}} \in \{0,1\}^{J \times J}$ is the binary
adjacency matrix, with $A_{\mathcal{S}}[i,j]=1$ if and only if joints $i$ and $j$ are
kinematic neighbors.
As shown in Fig.~\ref{fig:method}, our method \textsc{TAME} has
two components: a topology-aware
flow-matching~\citep{lipmanflow,liuflow,hu2023motion} transformer
with topology-constrained skeletal propagation (\textsc{TCSP}) that
serves as the editing backbone (Section~\ref{subsec:backbone}), and
\textsc{EFRA}, an edit-focused representation alignment strategy
(Section~\ref{subsec:train}). 


\paragraph{Motion representation.}
We represent a motion as a 3D tensor
$M \in \mathbb{R}^{T \times J \times D}$, where $T$ and $J$ denote
the frame and joint dimensions after padding for batching, and $D$
is the number of motion features per joint. For each joint $j$, its
feature vector $\phi_j \in \mathbb{R}^{12}$
is defined as
\[
\phi_j =
\begin{cases}
\{p_j, r_j, u_j\} \in \mathbb{R}^{12}, & j \neq j_r, \\
\{\Delta p_r, r_r, h_r\} \in \mathbb{R}^{12}, & j = j_r,
\end{cases}
\]
where $j_r$ denotes the root joint. For non-root joints,
$p_j \in \mathbb{R}^3$ is the root-relative position,
$r_j \in \mathbb{R}^6$ is the local joint rotation~\citep{zhou2019continuity}, and
$u_j \in \mathbb{R}^3$ is the velocity. For the root
joint, $\Delta p_r \in \mathbb{R}^3$ is the pelvis translation,
$r_r \in \mathbb{R}^6$ is the root orientation after removing the
heading rotation, and $h_r = (\sin\delta\theta, \cos\delta\theta,
z_{\text{root}}) \in \mathbb{R}^3$, where $\delta\theta$ is the
heading change and $z_{\text{root}}$ is the pelvis height.

\subsection{Topology-Aware Motion Editor Backbone}
\label{subsec:backbone}

Given the inputs $M_s$, $c$, $\mathcal{S}$ defined in
Section~\ref{subsec:overview} and a noised target motion
$\tilde{M}_t$, the flow-matching transformer of \textsc{TAME} predicts
the rectified-flow velocity
$\hat{v}_{\theta} \in \mathbb{R}^{T \times J \times D}$ that drives
the training objective in Section~\ref{subsec:train}. As illustrated
in Fig.~\ref{fig:method}, the architecture has two parts: an input
embedding stage, and a stack of $L$ identical \emph{\textsc{TAME} blocks}.

\paragraph{Input embedding.}
We embed each frame of the source motion and the noised target
motion as per-joint features. Since the non-root feature group
$\{p_j, r_j, u_j\}$ and the root feature group
$\{\Delta p_r, r_r, h_r\}$ have different semantics, we use separate
projections for root and non-root joints. Joint identity is provided
by a T5 text encoder applied to the joint names; for each joint
$j$, the resulting embedding $e_{\mathcal{S},j} \in \mathbb{R}^{d}$
is added to the joint token at every frame. To distinguish source
from target along the time axis and mark their boundary, we add
learnable source/target segment embeddings and insert a separator
token $M_{\mathrm{sep}} \in \mathbb{R}^{1 \times J \times d}$,
yielding
\[
\bar{M} = [\bar{M}_s; M_{\mathrm{sep}}; \bar{M}_t]
    \in \mathbb{R}^{(2T+1) \times J \times d}.
\]

\paragraph{\textsc{TAME} block with topology-constrained skeletal
propagation.}
Each \textsc{TAME} block contains three sequential attention
sub-layers operating on the same joint-time tokens along three
axes: joint, time, and text instruction. The joint-axis sub-layer
applies our topology-constrained skeletal propagation
(\textsc{TCSP}), in which self-attention is restricted to one-hop
kinematic neighbors. We achieve this by adding a mask
$H_{\mathcal{S}}$ to the pre-softmax attention scores, with $0$ for
neighbor pairs and $-\infty$ for non-neighbors, so that non-neighbor
attention is suppressed to zero after softmax:
\[
H_{\mathcal{S}}[i,j] =
\begin{cases}
0, & \text{if } i=j \text{ or } A_{\mathcal{S}}[i,j]=1, \\
-\infty, & \text{otherwise},
\end{cases}
\]
so that for each attention head,
\[
\operatorname{TCSP}(Q,K,V)
=
\operatorname{softmax}
\left(
\frac{QK^{\top}}{\sqrt{d_h}} + H_{\mathcal{S}}
\right)V,
\]
where $d_h$ is the head dimension. Information between distant
joints reaches them only through stacked layers along the kinematic
chain.

The time-axis sub-layer applies standard temporal self-attention,
allowing each target joint to attend to its corresponding
source-joint trajectory. The third sub-layer conditions motion
tokens on the text instruction via cross-attention, with motion
tokens as queries and CLIP-encoded text tokens as keys and values.
After $L$ blocks, we discard the source and separator segments,
apply a final normalisation, and use separate output heads for root
and non-root joints to produce
$\hat{v}_{\theta} \in \mathbb{R}^{T \times J \times D}$.

\subsection{Flow Matching with Edit-Focused Representation Alignment}
\label{subsec:train}
Our design starts from a key observation: motion editing is an
asymmetric learning problem. The unedited parts of the target
closely follow the source motion, which serves as a reliable
reference for them. The edited parts, in contrast, have to be
inferred from the source motion and the text instruction, and
therefore call for stronger guidance. We obtain this guidance from self-representation alignment
(SRA)~\citep{jiang2025no}, which builds on the observation that
the intermediate features of a diffusion transformer become more
informative as the noise decreases and the depth increases, and
therefore aligns the weaker features of the network to its better
ones without any external representation component. Accordingly,
\textsc{EFRA} aligns an earlier-layer student feature under higher
noise to a later-layer teacher feature under lower noise, where the
teacher is an EMA copy of the student (Fig.~\ref{fig:method},
upper right). Rather than applying the alignment to the whole
target motion, \textsc{EFRA} restricts it to the edit-relevant
joint-time tokens, identified by a DTW edit map derived from each
source--target pair (lower centre). The noise levels of the student
and the teacher are assigned by edit-aware dual-timestep
scheduling: the student takes a globally higher noise (lower left),
and the teacher takes the lower noise only on those tokens (lower
right).
Below we describe the three parts in turn: the DTW edit map, the
edit-aware dual-timestep scheduling, and the alignment objective.

\paragraph{DTW edit map.}
Restricting the alignment to the edited parts requires supervision
on where the edit takes place, which can be recovered by comparing
the source motion with the target motion. Prior
work~\citep{li2025simmotionedit,yang2025partmotionedit} derives
such supervision only at the frame or body-part level, too coarse
for joint-time tokens. To obtain a skeletal-temporally
structured edit signal, we use \emph{Dynamic Time Warping} (DTW)~\citep{sakoe1978dynamic} to
align source and target frames,
read out joint-level distances along the alignment, and threshold
them after per-sample min-max normalisation into a binary edit mask
$E \in \{0,1\}^{T \times J}$, where $E[i,j]=1$ marks the joint-time
tokens to be edited. The mask is precomputed offline for each source--target pair. The full construction
algorithm and per-step details are deferred to
Appendix~\ref{app:dtw}.

\paragraph{Edit-aware dual-timestep scheduling.}
To confine the teacher's guidance to the edited parts, we restrict
the information asymmetry between the teacher and the student to
the edit tokens. We realize this by decoupling the student's and
teacher's timesteps token-wise and injecting the edit mask $E$ into
the teacher's schedule, so that the teacher receives a lower noise
level than the student only on edit tokens. Given two
i.i.d.\ timesteps $t_1, t_2 \sim p(t)$ on $[0,1]$, we order them as
$t_{\mathrm{low}}=\min(t_1,t_2)$ and
$t_{\mathrm{high}}=\max(t_1,t_2)$, and define the token-wise teacher
timestep
\[
t_{\mathrm{teacher}}[i,j]=
\begin{cases}
t_{\mathrm{low}}, & \text{if } E[i,j]=1, \\
t_{\mathrm{high}}, & \text{otherwise}.
\end{cases}
\]
The student is paired with $t_{\mathrm{high}}$ uniformly across all
target tokens (Fig.~\ref{fig:method}, lower left), while the teacher
follows $t_{\mathrm{teacher}}$ (lower right). Under the
rectified-flow noise schedule with a shared noise
$\epsilon\sim\mathcal{N}(0,I)$, the noised inputs are
\[
\tilde{M}_t = (1-t_{\mathrm{high}})M_t + t_{\mathrm{high}}\epsilon,
\quad
\tilde{M}_t^{\mathrm{teacher}} =
(1-t_{\mathrm{teacher}})M_t + t_{\mathrm{teacher}}\epsilon,
\]
where $t_{\mathrm{teacher}}$ is broadcast element-wise across the
feature axis.

\paragraph{Self-distilled feature alignment.}
Finally, the alignment objective follows SRA in pairing an earlier
student layer $l_s$ with a later teacher layer $l_t$, so that the
teacher is ahead of the student in both noise level and processing
depth. The student and the teacher receive the same source motion,
text instruction, and skeleton, and differ only in their noised
input and their parameters: the teacher parameters $\theta'$ are an EMA
copy of the student parameters $\theta$, updated after every
optimiser step as $\theta' \leftarrow \alpha\theta' + (1-\alpha)\theta$
with momentum $\alpha$~\citep{jiang2025no}. For each target joint-time
token $(i,j)$, we take the output of the $l_s$-th block of the
student, $H^{l_s}_{\theta,i,j}\in\mathbb{R}^{d}$, and the output of
the $l_t$-th block of the teacher,
$H^{l_t}_{\theta',i,j}\in\mathbb{R}^{d}$. The student feature is
mapped by a two-layer MLP projection head $P_{\psi}$ and aligned to
the teacher feature, which is held fixed by the stop-gradient
operator $\operatorname{sg}(\cdot)$, with the cosine distance
$d_{\mathrm{cos}}$ over the edit tokens only
(Fig.~\ref{fig:method}, upper right):
\[
\mathcal{L}_{\mathrm{EFRA}}
=
\mathbb{E}
\left[
\frac{1}{|\Omega_E|}
\sum_{(i,j)\in\Omega_E}
d_{\mathrm{cos}}
\left(
P_{\psi}(H_{\theta,i,j}^{l_s}),
\operatorname{sg}(H_{\theta',i,j}^{l_t})
\right)
\right],
\]
where $\Omega_E=\{(i,j)\in\Omega\mid E[i,j]=1\}$ with $\Omega$ the
set of valid target joint-time tokens, and
$d_{\mathrm{cos}}(a,b)=1-\frac{a^\top b}{\|a\|_2\|b\|_2}$.
Combined with the standard rectified-flow objective on the student
velocity prediction $\hat{v}_{\theta}$ under the source, text, and
skeleton conditions, with target $v^{\ast}=\epsilon-M_t$,
\[
\mathcal{L}_{\mathrm{flow}}
=
\mathbb{E}
\left[
\frac{1}{|\Omega|}
\sum_{(i,j)\in\Omega}
\left\|
\hat{v}_{\theta,i,j}
-
v^{\ast}_{i,j}
\right\|_2^2
\right],
\]
we obtain the final training objective
\[
\mathcal{L}
=
\mathcal{L}_{\mathrm{flow}}
+
\lambda_{\mathrm{EFRA}}\mathcal{L}_{\mathrm{EFRA}},
\]
with weighting hyperparameter $\lambda_{\mathrm{EFRA}}$. The
alignment is fully self-distilled: the only added learnable
parameter is the projection head $P_{\psi}$, which is discarded at
inference. At inference, we deploy the EMA-teacher copy of the
backbone (rather than the student), so \textsc{EFRA} adds no
inference cost.

\section{Experiments}
\label{sec:experiments}

We start by introducing the multi-topology dataset
\textsc{TopoMotionFix} (Section~\ref{subsec:topomotionfix}),
together with our evaluation metrics and seen/unseen-topology
protocols (Section~\ref{subsec:exp_setup}). We then present the
main quantitative results, comparing \textsc{TAME} with
state-of-the-art motion-editing methods
(Section~\ref{subsec:main_results}). Next, we provide ablations on
the key components of \textsc{TAME}
(Section~\ref{subsec:ablations}). Finally, we demonstrate
qualitative results on both seen and unseen topologies
(Section~\ref{subsec:qualitative}).

\subsection{\textsc{TopoMotionFix} Dataset}
\label{subsec:topomotionfix}

\textsc{TopoMotionFix} comprises three components: textual edit instructions, source and target motion sequences, and the corresponding skeletal structures. 
Building on MotionFix~\citep{athanasiou2024motionfix}, we retarget its motion-skeleton pairs onto a pool of $83$ humanoid rigs sourced from LAFAN1, ACCAD, TotalCapture, PFNN, and SFU motion-capture datasets, augmented with skeleton randomisation following
\citep{lee2023same,liu2025text};
the full construction procedure is detailed in Appendix~\ref{app:dataset}.
\textsc{TopoMotionFix} contains $25{,}794$ distinct skeletal configurations, $40{,}304$ unique text-motion-skeleton
triplets, and $1{,}416$ unique textual edit instructions, inheriting MotionFix's train/test partition.

\subsection{Evaluation Setup}
\label{subsec:exp_setup}

\paragraph{Evaluation protocols.}
We define two evaluation protocols, both using the test pairs and the
evaluator provided by MotionFix~\citep{athanasiou2024motionfix}. In the
\emph{seen-topology} protocol, we train on MotionFix's training set
with only the input representation converted, enabling fair comparison
with prior fixed-topology motion-editing methods. In the
\emph{unseen-topology} protocol, the test set uses $10$ rigs, the
SMPL skeleton and $9$ held-out rigs, while the training set uses $83$
rigs. The held-out rigs are excluded from training and from the skeleton randomisation of
Algorithm~\ref{alg:skeleton_aug}, adapted from SAME~\citep{lee2023same},
so their topologies are fixed and unseen. The construction of the
held-out test sets and the evaluation of motions generated on non-SMPL
skeletons are detailed in Appendix~\ref{app:unseen}.

\paragraph{Evaluation metrics.}
We evaluate the semantic alignment of edited motions with both the
text instruction and the source motion.
Following MotionFix~\citep{athanasiou2024motionfix}, alignment is measured via motion-to-motion retrieval using pretrained TMR as a feature extractor.
%
%
We report R@1, R@2, R@3 and average rank (AvgR) on both generated-to-target ($g{\to}t$) and generated-to-source ($g{\to}s$) retrieval, with a fixed candidate batch of $32$ test-set motions (fixed across runs to eliminate randomness), and under the test-set setting (deferred to Appendix~\ref{app:fulltest}).
%

\paragraph{Baselines.}
For the seen-topology protocol, we compare \textsc{TAME} with current motion-editing methods, including
MDM~\citep{tevethuman}, MDM-BP and TMED~\citep{athanasiou2024motionfix},
MotionReFit~\citep{jiang2025dynamic},
SimMotionEdit~\citep{li2025simmotionedit},
OmniME~\citep{Shi_2026_CVPR}, and the cross-axis fusion model of
\citet{Han_2026_CVPR}. For the unseen-topology
protocol, no prior text-driven motion editing method operates on
arbitrary skeletons; to validate the effectiveness of our method, we
adapt AnyTop~\citep{gat2025anytop}, an arbitrary-topology motion
generation framework, by adapting its topological conditioning scheme
into our \textsc{TAME} editing pipeline.

\paragraph{Implementation details.}
\label{para:impl_details}
We adopt rectified flow matching~\citep{liuflow,lipmanflow} with $50$
sampling steps at inference on a single NVIDIA RTX A6000 ($48$\,GB)
GPU, $9$ \textsc{TAME} blocks, and a latent dimension of $512$. Each model is trained on $2$ NVIDIA RTX Pro $6000$ ($96$\,GB) GPUs
for approximately one day. More implementation details can be found in
Appendix~\ref{app:impl}. 

\subsection{Main Results}
\label{subsec:main_results}

\begin{table}[t]
\caption{Quantitative comparison. $\uparrow$
indicates higher values are better, $\downarrow$ indicates lower
values are better; bold indicates best; dash indicates
unavailable.}
\label{tab:batch32_results}
\centering
\scriptsize
\setlength{\tabcolsep}{2.2pt}
\resizebox{\linewidth}{!}{
\begin{tabular}{@{}lllcccccccc@{}}
\toprule
\multirow{2}{*}{Dataset} & \multirow{2}{*}{Protocol} & \multirow{2}{*}{Method}
& \multicolumn{4}{c}{Generated-to-Target (Batch-32)} &
\multicolumn{4}{c}{Generated-to-Source (Batch-32)} \\
\cmidrule(lr){4-7} \cmidrule(lr){8-11}
& &
& R@1$\uparrow$ & R@2$\uparrow$ & R@3$\uparrow$ & AvgR$\downarrow$
& R@1$\uparrow$ & R@2$\uparrow$ & R@3$\uparrow$ & AvgR$\downarrow$ \\
\midrule
 & & GT
& 100.00 & 100.00 & 100.00 & 1.00
& 74.01 & 84.52 & 89.91 & 2.03 \\
\midrule
MotionFix & Seen & MDM~\citep{tevethuman}
& 4.03 & 7.56 & 10.48 & 15.55
& 2.62 & 6.15 & 9.38 & 15.88 \\
MotionFix & Seen & MDM-BP~\citep{athanasiou2024motionfix}
& 39.10 & 50.09 & 54.84 & 6.46
& 61.28 & 69.55 & 73.99 & 4.21 \\
MotionFix & Seen & TMED~\citep{athanasiou2024motionfix}
& 64.11 & 76.51 & 82.56 & 2.72
& 73.19 & 85.08 & 91.03 & 1.92 \\
MotionFix & Seen & MotionReFit~\citep{jiang2025dynamic}
& 66.33 & 80.05 & 84.98 & 2.64
& 83.47 & 90.42 & 92.84 & 1.73 \\
MotionFix & Seen & SimMotionEdit~\citep{li2025simmotionedit}
& 70.62 & 82.92 & 88.12 & 2.38
& 72.71 & 83.54 & 87.50 & -- \\
MotionFix & Seen & \citet{Han_2026_CVPR}
& 74.38 & 88.54 & 92.08 & 1.92
& 78.96 & 91.04 & 93.33 & -- \\
MotionFix & Seen & OmniME~\citep{Shi_2026_CVPR}
& 77.29 & 88.54 & 91.88 & 1.79
& -- & -- & -- & -- \\
MotionFix & Seen & \textsc{TAME}
& \textbf{78.75} & \textbf{91.67} & \textbf{94.79} & \textbf{1.55}
& \textbf{86.04} & \textbf{94.58} & \textbf{97.50} & \textbf{1.36} \\
\midrule
\textsc{TopoMotionFix} & Unseen (SMPL+held-out rigs) & AnyTop~\citep{gat2025anytop}
& 68.85 & 83.87 & 89.72 & 2.01
& 84.17 & 92.54 & 95.26 & 1.42 \\
\textsc{TopoMotionFix} & Unseen (SMPL+held-out rigs) & \textsc{TAME} w/o \textsc{TCSP}, w/o \textsc{EFRA}
& 67.54 & 81.55 & 87.10 & 2.13
& 82.06 & 91.73 & 95.36 & 1.51 \\
\textsc{TopoMotionFix} & Unseen (SMPL+held-out rigs) & \textsc{TAME} w/o \textsc{EFRA}
& 71.67 & 85.48 & 91.23 & 1.97
& 87.00 & 93.75 & 96.88 & 1.39 \\
\textsc{TopoMotionFix} & Unseen (SMPL+held-out rigs) & \textsc{TAME}
& \textbf{72.78} & \textbf{86.69} & \textbf{91.43} & \textbf{1.95}
& \textbf{90.02} & \textbf{95.46} & \textbf{97.68} & \textbf{1.29} \\
\bottomrule
\end{tabular}
}
\end{table}

\paragraph{Quantitative comparisons.}
We report quantitative comparisons in
Table~\ref{tab:batch32_results}; test-set retrieval results are
reported in Table~\ref{tab:fulltest_results}
(Appendix~\ref{app:fulltest}). 
On the seen-topology setting, \textsc{TAME}
outperforms all fixed-topology baselines on both edit alignment and
source preservation.
We attribute this to edit-focused representation alignment (\textsc{EFRA}), which sharpens supervision on edit-relevant joint-time tokens via a dual-timestep self-distillation gap, and the per-joint motion representation, which preserves source content more faithfully than canonical pose vectors used by prior baselines. 
On the unseen-topology setting, \textsc{TAME} surpasses the AnyTop-adapted baseline by $+3.93$ batch R@1, demonstrating robustness to topologies unseen during training.

\subsection{Ablation Studies}
\label{subsec:ablations}

\textsc{TCSP}'s effectiveness for heterogeneous topologies is
established by Table~\ref{tab:batch32_results}. Here, in
Table~\ref{tab:component_ablation}, we further validate
\textsc{EFRA} and ablate its hyperparameters on seen-topology
MotionFix.

\begin{table}[t]
\caption{Ablations on MotionFix. The first two rows ablate
\textsc{TCSP} (\textsc{TAME} w/o \textsc{TCSP}, w/o \textsc{EFRA} vs.\
\textsc{TAME} w/o \textsc{EFRA}); blocks (a)-(c) keep \textsc{TCSP} and
\textsc{EFRA} enabled and vary one hyperparameter:
PH (projection-head dimension), Block (student/teacher layer pair
$(l_s, l_t)$), Edit Mask (\checkmark{}/$\times$: with/without the
DTW edit mask). Bold marks our default. $\uparrow$/$\downarrow$
indicates higher/lower is better.}
\label{tab:component_ablation}
\centering
\scriptsize
\setlength{\tabcolsep}{2.5pt}
\resizebox{\linewidth}{!}{
\begin{tabular}{@{}lccccccccccc@{}}
\toprule
& & & & \multicolumn{4}{c}{Generated-to-Target (Batch-32)} &
\multicolumn{4}{c}{Generated-to-Source (Batch-32)} \\
\cmidrule(lr){5-8} \cmidrule(lr){9-12}
& PH & Block & Edit Mask
& R@1$\uparrow$ & R@2$\uparrow$ & R@3$\uparrow$ & AvgR$\downarrow$
& R@1$\uparrow$ & R@2$\uparrow$ & R@3$\uparrow$ & AvgR$\downarrow$ \\
\midrule
\textsc{TAME} w/o \textsc{TCSP}, w/o \textsc{EFRA} & --- & --- & ---
& 74.80 & 88.00 & 92.64 & 1.80
& 84.38 & 92.64 & 95.16 & 1.52 \\
\textsc{TAME} w/o \textsc{EFRA} & --- & --- & ---
& 74.70 & 87.50 & 90.62 & 1.85
& 84.17 & 92.34 & 95.06 & 1.51 \\
\midrule
\emph{(a) Projection-head dimension} & 256 & 4/7 & \checkmark
& 76.11 & 89.11 & 93.45 & 1.66
& 84.27 & 92.84 & 94.86 & 1.53 \\
& 512 & 4/7 & \checkmark
& 77.32 & 88.31 & 92.74 & 1.74
& 84.48 & 92.44 & 94.76 & 1.59 \\
& \textbf{1024} & \textbf{4/7} & \textbf{\checkmark}
& \textbf{78.75} & \textbf{91.67} & \textbf{94.79} & \textbf{1.55}
& \textbf{86.04} & \textbf{94.58} & \textbf{97.50} & \textbf{1.36} \\
\midrule
\emph{(b) Alignment block layers $(l_s, l_t)$} & 1024 & 2/6 & \checkmark
& 76.01 & 89.01 & 93.15 & 1.70
& 83.47 & 92.24 & 94.66 & 1.61 \\
& 1024 & 3/7 & \checkmark
& 77.22 & 89.72 & 93.45 & 1.60
& 82.86 & 91.43 & 94.46 & 1.58 \\
& 1024 & 3/8 & \checkmark
& 76.21 & 88.21 & 92.94 & 1.65
& 81.35 & 91.63 & 94.35 & 1.60 \\
& \textbf{1024} & \textbf{4/7} & \textbf{\checkmark}
& \textbf{78.75} & \textbf{91.67} & \textbf{94.79} & \textbf{1.55}
& \textbf{86.04} & \textbf{94.58} & \textbf{97.50} & \textbf{1.36} \\
\midrule
\emph{(c) DTW edit mask} & \textbf{1024} & \textbf{4/7} & \textbf{\checkmark}
& \textbf{78.75} & \textbf{91.67} & \textbf{94.79} & \textbf{1.55}
& \textbf{86.04} & \textbf{94.58} & \textbf{97.50} & \textbf{1.36} \\
& 1024 & 4/7 & $\times$
& 75.71 & 87.70 & 92.54 & 1.74
& 82.06 & 91.33 & 94.56 & 1.64 \\
\bottomrule
\end{tabular}
}
\vspace{-0.3cm}
\end{table}

\paragraph{Effects of \textsc{TCSP} and \textsc{EFRA}.}
We ablate the components of \textsc{TAME} in
Table~\ref{tab:component_ablation}. The first two rows isolate
\textsc{TCSP}: on this single-topology setting, the gap between
\textsc{TAME} w/o \textsc{EFRA} and \textsc{TAME} w/o \textsc{TCSP},
w/o \textsc{EFRA} is only $-0.10$ edit R@1, indicating that the
adjacency prior is dispensable on a fixed skeleton. In contrast, in the unseen-topology setting
\textsc{TAME} surpasses AnyTop's content-dependent bias by
$+3.93$ edit R@1 (Table~\ref{tab:batch32_results}), indicating that
the hard topology constraint matters once the input topology varies.
\textsc{TAME} further surpasses \textsc{TAME} w/o \textsc{EFRA} by $+4.05$ edit
R@1, together with consistent gains on source preservation,
validating \textsc{EFRA} as the key design choice behind our
edit alignment.

\paragraph{\textsc{EFRA} hyperparameters.}
For the projection-head dimension, $d_p{=}256$ already lifts edit R@1 by $+1.41$ over \textsc{TAME} w/o \textsc{EFRA}, confirming that even a narrow alignment head injects useful self-distillation signal, and R@1 continues to climb through $d_p{=}1024$. 
For the student/teacher layer pair $(l_s, l_t)$, we sweep four configurations covering shallower SRA ratios and our deeper default; the mid-stack student to near-final teacher pair $(4, 7)$ is best, while pulling the student too early ($s{=}2$) or the teacher to the last layer ($t{=}8$) drops edit R@1 by ${\sim}2.5$--$2.7$, matching SRA's pattern. 
Removing the DTW edit mask reduces edit R@1 by $-3.04$ and source R@1 by $-3.98$: aligning over trivially preserved frames dilutes the learning signal on the tokens that actually carry the editing intent.

\subsection{Qualitative Results}
\label{subsec:qualitative}

\begin{figure*}[t]
\centering
\includegraphics[width=0.95\textwidth]{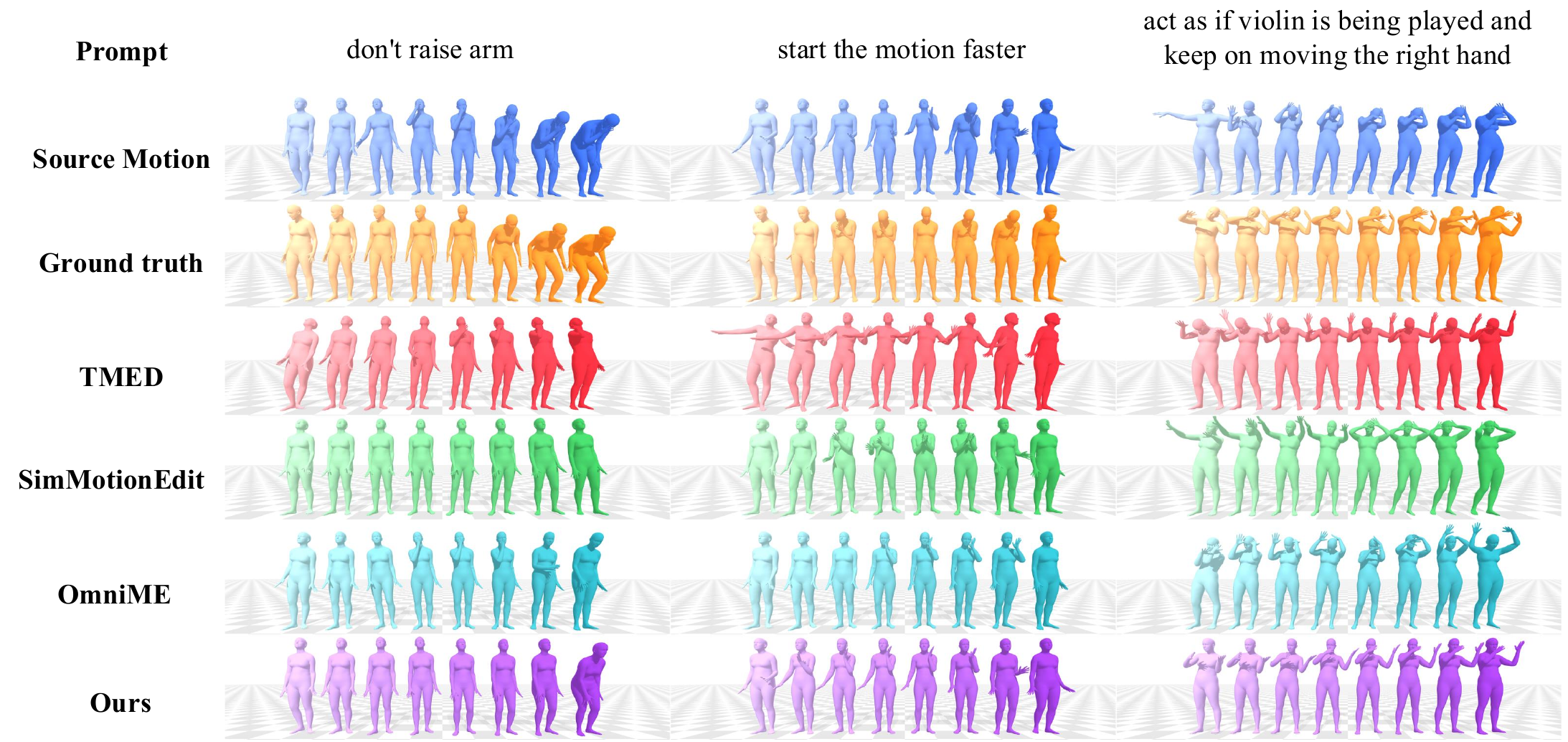}
\caption{Qualitative comparison on the seen-topology setting.
Columns are edit prompts; rows are source, ground truth,
TMED~\citep{athanasiou2024motionfix}, SimMotionEdit~\citep{li2025simmotionedit},
OmniME~\citep{Shi_2026_CVPR}, and our \textsc{TAME}. Each cell shows 8 keyframes (light to dark in time).}
\label{fig:qualitative}
\vspace{-0.3cm}
\end{figure*}

\textsc{TAME} avoids the editing failures that TMED, SimMotionEdit
and OmniME exhibit in Fig.~\ref{fig:qualitative}: ignoring
explicit prohibitions (column~1: TMED raises the arm in
keyframes~4--8 and OmniME in keyframes~4--6, while SimMotionEdit
obeys the instruction but loses the source's slow forward bend),
distorting the source action when applying speed edits (column~2:
TMED only copies the source endpoints, SimMotionEdit advances the
onset to keyframe~2 but misreads the single-arm raise as a
bilateral motion, and OmniME advances the onset only to keyframe~4
and lets the hand drift to the ear instead of the chest), and
deviating from the text semantics under complex or abstract
instructions (column~3: SimMotionEdit ignores the violin
instruction, TMED merely repeats the source's first half, and
OmniME raises both arms overhead instead of forming the violin
pose).

\begin{figure*}[t]
\centering
\includegraphics[width=0.95\textwidth]{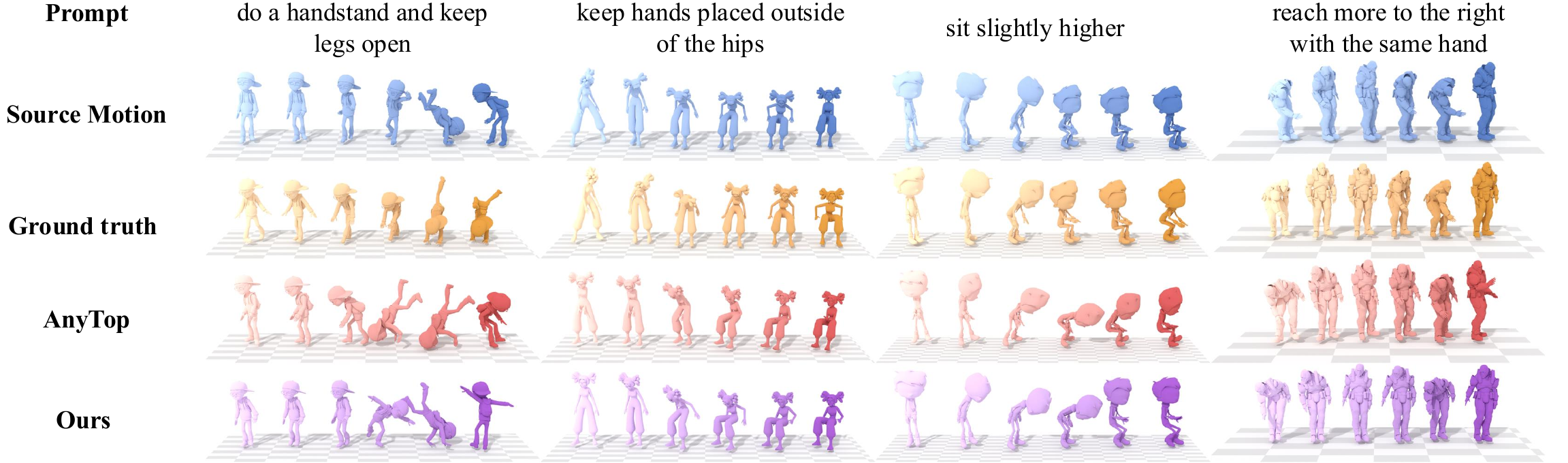}
\caption{Qualitative comparison on the unseen-topology setting,
evaluating \textsc{TopoMotionFix}-trained models on Mixamo
skeletons. Columns are edit prompts; rows are source, ground
truth, AnyTop~\citep{gat2025anytop}, and our \textsc{TAME}. Each
cell shows 6 keyframes (light to dark in time).}
\label{fig:qualitative_unseen}
\vspace{-0.3cm}
\end{figure*}

The unseen-topology examples in
Fig.~\ref{fig:qualitative_unseen} further demonstrate that
\textsc{TAME} does not simply memorize skeleton layouts from
training. When evaluated on Mixamo skeletons unseen during
training, our model still produces plausible local edits and
preserves the overall source motion trajectory. In contrast,
baselines more frequently introduce topology-dependent artifacts,
including unnatural limb trajectories, over-smoothed edits, or
unintended changes to unedited joints. These examples complement
the quantitative results by showing that the gains in retrieval
correspond to visually coherent, instruction-faithful edited
motions.

\section{Conclusion}
\label{sec:conclusion}

We studied text-driven motion editing across heterogeneous humanoid skeletons, a setting that better reflects the diverse character data encountered in real animation pipelines. To make the setting trainable and evaluable, we constructed \textsc{TopoMotionFix}, a multi-topology extension of MotionFix with seen- and unseen-topology evaluation protocols. 
We proposed \textsc{TAME}, a
topology-aware flow-matching transformer that represents motion as
joint-time tokens and uses topology-constrained skeletal propagation
(\textsc{TCSP}) to confine joint attention to one-hop kinematic
neighbors, so that the model operates natively on skeletons of
varying topology. Since the edited parts of a motion have to be
inferred from the source motion and the instruction and therefore
call for stronger guidance, we further introduced \textsc{EFRA}, a
self-distilled representation alignment strategy that aligns the
student to a cleaner EMA teacher only on the edit tokens identified
by a DTW edit map. On MotionFix, \textsc{TAME} outperforms
fixed-topology editing methods on both edit alignment and source
preservation. On \textsc{TopoMotionFix}, it generalizes robustly to
skeleton topologies unseen during training, bringing text-driven motion
editing to the varied characters of animation production.

%
%

\subsubsection*{AI use statement}
In this work, we did not use generative AI tools for tasks requiring mandatory
disclosure.
We did not use generative AI tools to develop theoretical models, formulate
mathematical claims, design experiments, implement methods, or interpret
experimental results, and the remaining mandatory-disclosure tasks are not
applicable to this work.
We used generative AI tools for code review, debugging, and minor code editing,
as well as to proofread the manuscript for grammar, clarity, and readability.
All AI-assisted work was reviewed by the authors.
AI-assisted code modifications were tested for correctness, and suggested
manuscript revisions were verified to preserve the technical meaning and
accuracy of the paper.
The authors take full responsibility for the final content of this work,
including all text and artifacts produced with the aid of generative AI.

\subsubsection*{Ethics statement}
Our perceptual study (Appendix~\ref{app:perceptual}) involved voluntary
adult participants who rated anonymised motion clips; no personal data was
collected. \textsc{TopoMotionFix} is derived from a publicly available motion
editing dataset and publicly released character rigs, and will be
released under the same terms as its sources. The method edits skeletal motion data
only and does not involve identifiable human appearance.

\subsubsection*{Reproducibility statement}
Implementation details of \textsc{TAME}, including the architecture, training
configuration, and all \textsc{EFRA} hyperparameters, are given in
Appendix~\ref{app:impl}. The \textsc{TopoMotionFix} construction procedure and
the DTW edit-map construction are specified with pseudocode in
Appendices~\ref{app:dataset} and~\ref{app:dtw}. The evaluation protocols and
metrics follow the public MotionFix test set and evaluator, as described in
Section~\ref{subsec:exp_setup} and Appendix~\ref{app:fulltest}. The
unseen-topology protocol, including the construction of the held-out
test set and the evaluation of motions generated on non-SMPL skeletons,
is detailed in Appendix~\ref{app:unseen}.

\bibliography{references}
\bibliographystyle{iclr2027_conference}

\newpage
\input{appendix}

\end{document}

%% file: math_commands.tex
\usepackage{amsmath,amsfonts,bm}

\def\eqref#1{equation~\ref{#1}}

\def\1{\bm{1}}

\DeclareMathAlphabet{\mathsfit}{\encodingdefault}{\sfdefault}{m}{sl}
\SetMathAlphabet{\mathsfit}{bold}{\encodingdefault}{\sfdefault}{bx}{n}



%% file: appendix.tex
\appendix

\section{\textsc{TopoMotionFix} Construction Details}
\label{app:dataset}

This appendix details the \textsc{TopoMotionFix} construction
summarised in Section~\ref{subsec:topomotionfix}. We retarget
each MotionFix editing pair onto a set of sampled skeletons while
keeping the original text instruction unchanged. The retargeting
pipeline largely reuses the protocol of SAME~\citep{lee2023same},
which is also adopted by SkeleMotion-3D~\citep{liu2025text}.

The full pipeline is split into two stages.
Algorithm~\ref{alg:skeleton_aug} defines the per-skeleton
augmentation routine: starting from a base skeleton sampled from a
character pool, we randomly add or remove joints around the spine
and neck, optionally insert dummy joints at the hip, lower back,
and shoulder, zero out random end-effector offsets, scale per-axis
bone offsets uniformly with $[s_{\min},s_{\max}]=[0.7,1.3]$, apply left-right
symmetric scaling, and randomly sample a global height. Algorithm~\ref{alg:topomotionfix_construction}
drives the overall dataset construction: for each MotionFix editing
pair, we first append the original SMPL pair, then generate
augmented skeletons via Algorithm~\ref{alg:skeleton_aug}, retarget
both source and target motions onto each augmented skeleton (the
editing pair must share the same target skeleton), and convert each
retargeted clip into our per-joint motion representation.

The resulting dataset packages the $6{,}730$ source-target editing
pairs into $40{,}304$ editing instances ($6{,}730$ on the original
SMPL skeleton plus $33{,}574$ on randomised humanoid skeletons),
spanning $25{,}794$ unique skeletal configurations. Each instance
carries one source motion and one target motion on a shared
skeleton, giving $80{,}608$ motion-skeleton clips
at the per-motion level. The $6{,}730$
textual editing instructions are inherited from MotionFix and
contain $1{,}416$ unique sentences across $56{,}728$ total words,
averaging $8.4$ words per instruction.

\begin{algorithm}[t]
\footnotesize
\caption{Skeleton augmentation (adapted from~\citep{lee2023same})}
\label{alg:skeleton_aug}
\KwIn{Character skeleton pool $\mathcal{C}$;\ \ bone-scale range $[s_{\min},s_{\max}]$;\ \ height-scale range $[h_{\min},h_{\max}]$}
\KwOut{Augmented skeleton $\mathcal{S}'$}

\textbf{1. Sample base skeleton.}~\\
$\mathcal{S}\sim\mathrm{Uniform}(\mathcal{C})$;\quad $\mathcal{S}'\leftarrow\mathrm{copy}(\mathcal{S})$\;

\textbf{2. Joint-level perturbations.}~\\
\For{joint $j\in\mathcal{S}'.\mathrm{joints}$}{
    \If{$j\in\{\mathrm{Hip},\mathrm{LowerBack},\mathrm{Shoulder}\}$}{
        randomly insert dummy joints adjacent to $j$\;
    }
    \If{$j\in\{\mathrm{Spine},\mathrm{Neck}\}$}{
        randomly add or remove sub-joints along $j$'s chain\;
    }
    \If{$j\in\{\mathrm{EndEffectors}\}$}{
        $j.\mathrm{offset}\leftarrow\mathbf{0}$ with random probability\;
    }
    sample $s_x,s_y,s_z\overset{\mathrm{iid}}{\sim}\mathcal{U}(s_{\min},s_{\max})$;\ $j.\mathrm{offset}\leftarrow j.\mathrm{offset}\odot(s_x,s_y,s_z)$\;
}

\textbf{3. Global scaling.}~\\
apply left-right symmetric bone-offset scaling\;
sample $h\sim\mathcal{U}(h_{\min},h_{\max})$;\ \ $\mathcal{S}'.\mathrm{height}\leftarrow\mathcal{S}'.\mathrm{height}\cdot h$\;

\Return $\mathcal{S}'$\;
\end{algorithm}

\begin{algorithm}[t]
\footnotesize
\caption{\textsc{TopoMotionFix} construction}
\label{alg:topomotionfix_construction}
\KwIn{MotionFix editing pairs $\mathcal{P}=\{(M_s^n,M_t^n,c^n)\}_{n=1}^{N}$ on the original SMPL skeleton $\mathcal{S}^n_{0}$ \\
character skeleton pool $\mathcal{C}$ \\
augmentations per pair $K$}
\KwOut{Multi-topology editing dataset $\mathcal{D}_{\mathrm{topo}}$}

\textbf{1. Initialise.}~\\
$\mathcal{D}_{\mathrm{topo}}\leftarrow\emptyset$\;

\textbf{2. Iterate over editing pairs.}~\\
\For{each pair $(M_s^n,M_t^n,c^n)\in\mathcal{P}$}{
    \textbf{2.1 Original SMPL pass:}~\\
    $\mathcal{D}_{\mathrm{topo}}\leftarrow\mathcal{D}_{\mathrm{topo}}\cup\{(M_s^n,M_t^n,c^n,\mathcal{S}^n_{0})\}$\;

    \textbf{2.2 Augmented topology pass:}~\\
    \For{$k=1$ to $K$}{
        $\mathcal{S}^{n,k}\leftarrow\textsc{SkeletonAugment}(\mathcal{C})$\tcp*{Algorithm~\ref{alg:skeleton_aug}}
        \For{$r\in\{\mathrm{src},\mathrm{tgt}\}$}{
            $M_r^{n,k}\leftarrow\textsc{Retarget}(M_r^n,\mathcal{S}^{n,k})$\;
        }
        $\mathcal{D}_{\mathrm{topo}}\leftarrow\mathcal{D}_{\mathrm{topo}}\cup\{(M_s^{n,k},M_t^{n,k},c^n,\mathcal{S}^{n,k})\}$\;
    }
}

\Return $\mathcal{D}_{\mathrm{topo}}$\;
\end{algorithm}

\section{Unseen-Topology Protocol and Evaluation on Non-SMPL Skeletons}
\label{app:unseen}

This appendix details the unseen-topology protocol summarised in
Section~\ref{subsec:exp_setup}: how the held-out test sets are built
and how motions generated on non-SMPL skeletons are scored with the
MotionFix evaluator.

\paragraph{Held-out rigs and test sets.}
\textsc{TopoMotionFix} contains $92$ humanoid rigs in total, of
which $9$ are assigned to the test set and $83$ to the training set.
Together with the SMPL skeleton of MotionFix, the $9$ held-out rigs
make up the $10$ test topologies. Each held-out rig is used only in the
test set, with its original, fixed topology and bone lengths. The held-out rigs are removed both from the training samples
and from the character pool $\mathcal{C}$ from which
Algorithm~\ref{alg:skeleton_aug} samples base skeletons, so no
randomised variant of a held-out rig appears in training. The
samples of the unseen-topology test set come from two sources. The
first is the original MotionFix test split of $1{,}013$ source--target
pairs on the SMPL skeleton. The second is built by retargeting each of
these $1{,}013$ pairs, with the retargeting pipeline of
SAME~\citep{lee2023same}, onto $4$ rigs drawn at random from the $9$
non-SMPL held-out rigs. The unseen-topology test set therefore contains
$5{,}065$ source--target--instruction triplets: $1{,}013$ on the SMPL
skeleton and $4{,}052$ on the $9$ held-out rigs.

\paragraph{Evaluation on non-SMPL skeletons.}
The evaluator provided by MotionFix is a TMR
model~\citep{Petrovich_2023_ICCV} trained on the $135$-dimensional SMPL
representation (pelvis translation, $6$D pelvis orientation, and $6$D
rotations of the $21$ SMPL body joints). We retain this evaluator rather
than training a topology-specific one, since it underlies every
published result we compare against and thus keeps the
unseen-topology scores directly comparable with their seen-topology
counterparts. Motions
generated on non-SMPL rigs are therefore retargeted onto the SMPL
skeleton before scoring. Each edited motion generated on a non-SMPL
skeleton is retargeted with the same SAME pipeline, converted to the
$135$-dimensional representation, and scored with the retrieval
metrics. For scoring, the $4{,}052$ retargeted triplets are divided
into four subsets of $1{,}013$, each containing every test pair
exactly once. Each subset of $1{,}013$ samples is then evaluated
exactly as in the single-topology setting, under both the batch-$32$
and the test-set gallery settings. The unseen-topology
scores in Tables~\ref{tab:batch32_results} and~\ref{tab:fulltest_results}
are the average over the five subsets, namely the SMPL subset and the
four retargeted subsets. Since the retargeting from a rig back to SMPL is itself
lossy, we additionally report a ground-truth ceiling in
Table~\ref{tab:fulltest_results} (GT (retargeted)), obtained by
retargeting the MotionFix test targets onto the held-out rigs and back
to SMPL through the same pipeline. The round trip changes every metric
by at most $0.02$ points relative to the original ground truth, so the
retargeting loss is acceptable.

\section{DTW Edit Map Construction}
\label{app:dtw}

This appendix details the DTW edit map construction summarised
in Section~\ref{subsec:train}. The mask
$E \in \{0,1\}^{T \times J}$ is constructed for each source--target
pair and stored with the dataset. We write
$p_{i,j}\in\mathbb{R}^3$ for the position and
$r_{i,j}\in\mathbb{R}^6$ for the $6$D rotation of joint $j$ at
frame $i$, with superscripts $s$ and $t$ marking the source and
edited target.

The full procedure is given in Algorithm~\ref{alg:dtw_edit_map}: we first
compute a joint-level frame-pair distance from the position and
rotation components with equal weights, then assemble a frame
cost matrix where pairs beyond a window of width $b$ on the
diagonal are masked to $+\infty$, and run classical DTW on this matrix to
obtain the optimal source-to-target alignment path. The path
identifies, for each source frame, its best-aligned target frame;
we extract the per-joint distance at this target frame, apply
per-sample min-max normalisation into $[0,1]$, and threshold with
$\tau_g$ to obtain the binary mask.

This window absorbs small temporal misalignments between source
and target while preventing pathological matches between distant
parts of the sequence; per-sample min-max normalisation makes
the threshold
$\tau_g$ comparable across samples regardless of their overall
edit intensity. Throughout, we use $b=10$ frames and
$\tau_g=0.2$.

\begin{algorithm}[t]
\footnotesize
\caption{DTW edit map construction}
\label{alg:dtw_edit_map}
\KwIn{Source motion $M_s$ and edited target motion $M_t$, both in $\mathbb{R}^{T\times J\times D}$;\ \ threshold $\tau_g$;\ \ window width $b$}
\KwOut{Binary edit mask $E\in\{0,1\}^{T\times J}$}

\textbf{1. Joint-level frame-pair distance.}~\\
$d_{i,i',j}\leftarrow\tfrac{1}{2}\|p^s_{i,j}-p^t_{i',j}\|_1+\tfrac{1}{2}\|r^s_{i,j}-r^t_{i',j}\|_1$\;

\textbf{2. Windowed frame cost matrix.}~\\
$C_{i,i'}\leftarrow\begin{cases}
\tfrac{1}{J}\sum_{j=1}^{J} d_{i,i',j}, & |i-i'|\leq b\\
+\infty, & \text{otherwise}
\end{cases}$\;

\textbf{3. DTW alignment.}~\\
$\Pi\leftarrow\mathrm{DTW}(C)\subseteq[T]\times[T]$\tcp*{optimal alignment path}
$\pi(i)\leftarrow\arg\min_{i':(i,i')\in\Pi} C_{i,i'}$\tcp*{best-aligned target frame}
$\bar{d}_{i,j}\leftarrow d_{i,\pi(i),j}$\tcp*{per-joint distance along path}

\textbf{4. Per-sample min-max normalisation.}~\\
$G_{i,j}\leftarrow(\bar{d}_{i,j}-\min_{i,j}\bar{d}_{i,j})/(\max_{i,j}\bar{d}_{i,j}-\min_{i,j}\bar{d}_{i,j})$\;

\textbf{5. Threshold to binary mask.}~\\
$E_{i,j}\leftarrow\mathbf{1}[G_{i,j}>\tau_g]$\;

\Return $E$\;
\end{algorithm}

\section{Test-Set Retrieval Results}
\label{app:fulltest}

Following the official evaluation implementation of
SimMotionEdit~\citep{li2025simmotionedit}, which is also adopted by
\citet{Han_2026_CVPR} and OmniME~\citep{Shi_2026_CVPR},
we additionally report the retrieval metrics of
Section~\ref{subsec:exp_setup} under its test-set setting, in which
each generated motion is ranked among the test motions rather than
within a batch of $32$. Table~\ref{tab:fulltest_results} lists the
results; baseline numbers are taken from the respective papers, and
a dash indicates an unavailable number. Under both
protocols, \textsc{TAME} obtains the best score on every metric, for
edit alignment and source preservation alike.

\begin{table}[t]
\caption{Quantitative comparison under the test-set retrieval
setting of the official SimMotionEdit evaluation implementation,
with the same metrics as Table~\ref{tab:batch32_results}. Baseline
numbers are taken from the respective papers; dash indicates
unavailable; bold indicates best within each protocol.}
\label{tab:fulltest_results}
\centering
\scriptsize
\setlength{\tabcolsep}{2.2pt}
\resizebox{\linewidth}{!}{
\begin{tabular}{@{}lllcccccccc@{}}
\toprule
\multirow{2}{*}{Dataset} & \multirow{2}{*}{Protocol} & \multirow{2}{*}{Method}
& \multicolumn{4}{c}{Generated-to-Target (Test set)} &
\multicolumn{4}{c}{Generated-to-Source (Test set)} \\
\cmidrule(lr){4-7} \cmidrule(lr){8-11}
& &
& R@1$\uparrow$ & R@2$\uparrow$ & R@3$\uparrow$ & AvgR$\downarrow$
& R@1$\uparrow$ & R@2$\uparrow$ & R@3$\uparrow$ & AvgR$\downarrow$ \\
\midrule
 & & GT
& 64.36 & 88.75 & 95.56 & 1.74
& 20.83 & 33.66 & 40.47 & 33.13 \\
 & & GT (retargeted)
& 64.34 & 88.73 & 95.55 & 1.73
& 20.82 & 33.65 & 40.47 & 33.12 \\
\midrule
MotionFix & Seen & MDM~\citep{tevethuman}
& 0.00 & 0.30 & 0.49 & 476.65
& 0.10 & 0.10 & 0.10 & 486.26 \\
MotionFix & Seen & MDM-BP~\citep{athanasiou2024motionfix}
& 8.69 & 14.71 & 18.36 & 180.99
& 30.21 & 36.82 & 40.47 & 106.05 \\
MotionFix & Seen & TMED~\citep{athanasiou2024motionfix}
& 14.51 & 21.72 & 28.73 & 56.63
& 22.41 & 34.45 & 40.57 & 31.42 \\
MotionFix & Seen & MotionReFit~\citep{jiang2025dynamic}
& 14.13 & 23.52 & 30.53 & 54.06
& 43.77 & 56.72 & 64.13 & 24.09 \\
MotionFix & Seen & SimMotionEdit~\citep{li2025simmotionedit}
& 25.49 & 39.33 & 49.21 & 23.49
& -- & -- & -- & -- \\
MotionFix & Seen & \citet{Han_2026_CVPR}
& 29.45 & 45.26 & 54.55 & 16.42
& -- & -- & -- & -- \\
MotionFix & Seen & OmniME~\citep{Shi_2026_CVPR}
& 32.02 & 50.20 & 59.88 & 13.06
& -- & -- & -- & -- \\
MotionFix & Seen & \textsc{TAME}
& \textbf{34.98} & \textbf{50.40} & \textbf{60.08} & \textbf{10.33}
& \textbf{54.35} & \textbf{66.60} & \textbf{73.72} & \textbf{7.58} \\
\midrule
\textsc{TopoMotionFix} & Unseen (SMPL+held-out rigs) & AnyTop~\citep{gat2025anytop}
& 24.70 & 37.75 & 45.26 & 17.49
& 46.25 & 60.47 & 67.59 & 8.85 \\
\textsc{TopoMotionFix} & Unseen (SMPL+held-out rigs) & \textsc{TAME} w/o \textsc{TCSP}, w/o \textsc{EFRA}
& 23.72 & 37.15 & 45.85 & 20.14
& 44.07 & 57.51 & 65.81 & 7.82 \\
\textsc{TopoMotionFix} & Unseen (SMPL+held-out rigs) & \textsc{TAME} w/o \textsc{EFRA}
& 26.68 & \textbf{42.09} & 50.79 & 17.11
& 54.74 & 66.80 & 72.13 & 8.80 \\
\textsc{TopoMotionFix} & Unseen (SMPL+held-out rigs) & \textsc{TAME}
& \textbf{28.46} & \textbf{42.09} & \textbf{53.16} & \textbf{16.53}
& \textbf{63.83} & \textbf{79.05} & \textbf{83.60} & \textbf{6.96} \\
\bottomrule
\end{tabular}
}
\end{table}

\section{Implementation Details}
\label{app:impl}

We adopt rectified flow matching~\citep{liuflow,lipmanflow}
with the linear interpolation of Section~\ref{subsec:train}. The
training timestep $t \in [0,1]$ is drawn uniformly at random from a
grid on $[0,1]$ with spacing $1/1{,}000$, and inference uses $50$
Euler sampling steps. Following
\citet{athanasiou2024motionfix}, we apply two-way classifier-free
guidance with a guidance scale of $2.5$ for the text condition
and $2.0$ for the source motion condition, with dropout $0.05$
applied independently on each of the text, source-motion, and
joint conditions during training. We encode the natural-language edit instruction with a frozen
CLIP-ViT-L/14 text encoder~\citep{radford2021learning} (injected
through cross-attention), and the joint-name strings of the
target skeleton with the pretrained T5-base text
encoder~\citep{raffel2020exploring} (added to the per-joint
token embeddings).

The flow-matching transformer consists of $L=9$ \textsc{TAME}
blocks with $8$ attention heads, a latent dimension of $512$,
and an FFN dimension of $1024$. Our edit-focused representation
alignment (\textsc{EFRA}) uses an EMA teacher with momentum
$\alpha=0.9999$, a projection head with hidden dimension $1024$,
and an alignment loss connecting student layer $4$ to teacher
layer $7$ via cosine distance restricted to the DTW-derived edit
mask (threshold $\tau_g=0.2$), with auxiliary loss weight
$\lambda_{\mathrm{EFRA}}=0.5$ relative to the rectified-flow
objective.

We train using the AdamW
optimiser~\citep{loshchilov2019adamw} with an initial learning
rate of $1{\times}10^{-4}$ decayed to $1{\times}10^{-6}$ along a
cosine schedule with $300$ warmup steps, weight decay $0.01$, for
$1{,}000$ epochs at an effective batch size of $128$, taking
approximately one day on two NVIDIA RTX Pro $6000$ ($96$\,GB)
GPUs. We implement \textsc{TAME} in PyTorch Lightning and train
every ablation from scratch with this configuration unless
otherwise noted.

\section{Perceptual Study}
\label{app:perceptual}

To complement the retrieval-based metrics in the main paper, we
conduct a perceptual study with human participants to assess how
well \textsc{TAME}'s edits align with text instructions and how
plausible the resulting motions appear, compared with
SimMotionEdit under the seen-topology protocol and AnyTop under
the unseen-topology protocol.

\subsection{Setup}
\label{app:perceptual:setup}

We run two parallel studies, one for the seen-topology protocol
and one for the unseen-topology protocol of
Table~\ref{tab:batch32_results}. Each participant in a study
observes $10$ sample sets, randomly drawn from the MotionFix test
split. Within each sample set, the participant is shown the
source motion, the text edit instruction, and three candidate
edited motions:
\begin{itemize}
\item \textbf{Seen-topology study}: ground truth (GT),
\textsc{TAME}, and
SimMotionEdit~\citep{li2025simmotionedit}.
\item \textbf{Unseen-topology study}: ground truth (GT),
\textsc{TAME}, and AnyTop~\citep{gat2025anytop}.
\end{itemize}
The three candidates are presented in random order within each
sample set to avoid positional bias. Participants rate each
candidate on two metrics, both on a $3$-point Likert scale
(higher is better):
\begin{itemize}
\item \emph{Alignment with text}: how well the edited motion
follows the instruction, irrespective of motion quality.
$1$ = the edited motion is completely different from the
instruction; $2$ = the edited motion follows only some parts
of the instruction; $3$ = the edited motion fully follows the
instruction.
\item \emph{Plausibility}: the quality of the edited motion,
irrespective of instruction-following. $1$ = severe issues
(self-intersections, unrealistic poses or movements);
$2$ = minor issues (foot sliding, plausible but slightly
non-humanlike movements); $3$ = no issues.
\end{itemize}
We follow the protocol of SimMotionEdit~\citep{li2025simmotionedit}
for the rating scales to allow direct comparison with their
reported numbers under the seen-topology protocol.

\subsection{Results}
\label{app:perceptual:results}

We recruited $N=25$ participants per study, consisting of
graduate students and staff from a university campus, yielding
$250$ ratings per candidate per metric in each study.
Table~\ref{tab:perceptual} reports the mean Likert score and
standard deviation aggregated across all participants and sample
sets, for the seen-topology and unseen-topology protocols.

\begin{table}[!htbp]
\caption{Perceptual evaluation. Mean $\pm$ standard deviation on
the $3$-point Likert scale across $25$ participants and $10$
sample sets per participant ($250$ ratings per cell). Higher is
better. Each block compares ground truth (GT), our \textsc{TAME},
and one baseline: SimMotionEdit for the seen-topology protocol
and AnyTop for the unseen-topology protocol.}
\label{tab:perceptual}
\centering
\small
\begin{tabular}{@{}llcc@{}}
\toprule
Protocol & Method & Alignment $\uparrow$ & Plausibility $\uparrow$ \\
\midrule
\multirow{3}{*}{Seen}
 & GT & 2.53 $\pm$ 0.66 & 2.61 $\pm$ 0.69 \\
 & SimMotionEdit~\citep{li2025simmotionedit} & 2.08 $\pm$ 0.81 & 2.28 $\pm$ 0.76 \\
 & \textsc{TAME} (ours) & \textbf{2.31 $\pm$ 0.74} & \textbf{2.46 $\pm$ 0.71} \\
\midrule
\multirow{3}{*}{Unseen}
 & GT & 2.53 $\pm$ 0.66 & 2.61 $\pm$ 0.69 \\
 & AnyTop~\citep{gat2025anytop} & 1.73 $\pm$ 0.84 & 1.98 $\pm$ 0.81 \\
 & \textsc{TAME} (ours) & \textbf{1.96 $\pm$ 0.79} & \textbf{2.14 $\pm$ 0.77} \\
\bottomrule
\end{tabular}
\end{table}

Across both protocols, participants prefer \textsc{TAME} over
the compared baseline by approximately $0.2$ Likert points on
both metrics, mirroring the retrieval-based gaps in the main
paper. The Alignment gap is slightly larger than the Plausibility
gap, indicating that human observers recognise \textsc{TAME}'s
edits as more faithful to the instruction while the baseline
motions can still appear visually clean. Comparing \textsc{TAME}
across protocols, scores drop by $0.32$--$0.35$ points moving
from seen to unseen topology, reflecting the additional
difficulty of editing motions on skeletons that the model has not
seen during training; even so, \textsc{TAME} remains substantially
ahead of AnyTop, which is not designed for text-based editing.

%% file: references.bib
@article{jiang2025no,
  title={No other representation component is needed: Diffusion transformers can provide representation guidance by themselves},
  author={Jiang, Dengyang and Wang, Mengmeng and Li, Liuzhuozheng and Zhang, Lei and Wang, Haoyu and Wei, Wei and Dai, Guang and Zhang, Yanning and Wang, Jingdong},
  journal={arXiv preprint arXiv:2505.02831},
  year={2025}
}

@inproceedings{athanasiou2024motionfix,
  title={Motionfix: Text-driven 3d human motion editing},
  author={Athanasiou, Nikos and Cseke, Alp{\'a}r and Diomataris, Markos and Black, Michael J and Varol, G{\"u}l},
  booktitle={SIGGRAPH Asia 2024 Conference Papers},
  pages={1--11},
  year={2024}
}

@inproceedings{li2025simmotionedit,
  title={SimMotionEdit: Text-Based Human Motion Editing with Motion Similarity Prediction},
  author={Li, Zhengyuan and Cheng, Kai and Ghosh, Anindita and Bhattacharya, Uttaran and Gui, Liangyan and Bera, Aniket},
  booktitle={Proceedings of the IEEE/CVF Conference on Computer Vision and Pattern Recognition},
  pages={27827--27837},
  year={2025}
}

@inproceedings{delmas2023posefix,
  title={Posefix: Correcting 3d human poses with natural language},
  author={Delmas, Ginger and Weinzaepfel, Philippe and Moreno-Noguer, Francesc and Rogez, Gr{\'e}gory},
  booktitle={Proceedings of the IEEE/CVF International Conference on Computer Vision},
  pages={15018--15028},
  year={2023}
}

@article{motionclr,
  title={MotionCLR: Motion Generation and Training-free Editing via Understanding Attention Mechanisms},
  author={Chen, Ling-Hao and Dai, Wenxun and Ju, Xuan and Lu, Shunlin and Zhang, Lei},
  journal={arxiv:2410.18977},
  year={2024}
}

@article{yang2025partmotionedit,
  title={PartMotionEdit: Fine-Grained Text-Driven 3D Human Motion Editing via Part-Level Modulation},
  author={Yang, Yujie and Zhang, Zhichao and Chen, Jiazhou and Wu, Zichao},
  journal={arXiv preprint arXiv:2512.24200},
  year={2025}
}

@inproceedings{gat2025anytop,
  title={Anytop: Character animation diffusion with any topology},
  author={Gat, Inbar and Raab, Sigal and Tevet, Guy and Reshef, Yuval and Bermano, Amit Haim and Cohen-Or, Daniel},
  booktitle={Proceedings of the Special Interest Group on Computer Graphics and Interactive Techniques Conference Conference Papers},
  pages={1--10},
  year={2025}
}

@inproceedings{lee2023same,
  title={Same: Skeleton-agnostic motion embedding for character animation},
  author={Lee, Sunmin and Kang, Taeho and Park, Jungnam and Lee, Jehee and Won, Jungdam},
  booktitle={SIGGRAPH Asia 2023 Conference Papers},
  pages={1--11},
  year={2023}
}

@inproceedings{liu2025text,
  title={Text-to-Any-Skeleton Motion Generation Without Retargeting},
  author={Liu, Qingyuan and Lv, Ke and Dong, Kun and Xue, Jian and Niu, Zehai and Wang, Jinbao},
  booktitle={Proceedings of the IEEE/CVF International Conference on Computer Vision},
  pages={12926--12936},
  year={2025}
}

@inproceedings{yang2024omnimotiongpt,
  title={Omnimotiongpt: Animal motion generation with limited data},
  author={Yang, Zhangsihao and Zhou, Mingyuan and Shan, Mengyi and Wen, Bingbing and Xuan, Ziwei and Hill, Mitch and Bai, Junjie and Qi, Guo-Jun and Wang, Yalin},
  booktitle={Proceedings of the IEEE/CVF Conference on Computer Vision and Pattern Recognition},
  pages={1249--1259},
  year={2024}
}

@inproceedings{tevethuman,
  title={Human Motion Diffusion Model},
  author={Tevet, Guy and Raab, Sigal and Gordon, Brian and Shafir, Yoni and Cohen-Or, Daniel and Bermano, Amit Haim},
  booktitle={The Eleventh International Conference on Learning Representations},
  year={2023}
}

@inproceedings{zhou2019continuity,
  title={On the continuity of rotation representations in neural networks},
  author={Zhou, Yi and Barnes, Connelly and Lu, Jingwan and Yang, Jimei and Li, Hao},
  booktitle={Proceedings of the IEEE/CVF conference on computer vision and pattern recognition},
  pages={5745--5753},
  year={2019}
}

@article{zhang2024motiondiffuse,
  title={Motiondiffuse: Text-driven human motion generation with diffusion model},
  author={Zhang, Mingyuan and Cai, Zhongang and Pan, Liang and Hong, Fangzhou and Guo, Xinying and Yang, Lei and Liu, Ziwei},
  journal={IEEE transactions on pattern analysis and machine intelligence},
  volume={46},
  number={6},
  pages={4115--4128},
  year={2024},
  publisher={IEEE}
}

@inproceedings{guo2024momask,
  title={Momask: Generative masked modeling of 3d human motions},
  author={Guo, Chuan and Mu, Yuxuan and Javed, Muhammad Gohar and Wang, Sen and Cheng, Li},
  booktitle={Proceedings of the IEEE/CVF Conference on Computer Vision and Pattern Recognition},
  pages={1900--1910},
  year={2024}
}

@inproceedings{jiang2025dynamic,
  title={Dynamic motion blending for versatile motion editing},
  author={Jiang, Nan and Li, Hongjie and Yuan, Ziye and He, Zimo and Chen, Yixin and Liu, Tengyu and Zhu, Yixin and Huang, Siyuan},
  booktitle={Proceedings of the Computer Vision and Pattern Recognition Conference},
  pages={22735--22745},
  year={2025}
}

@inproceedings{yurepresentation,
  title={Representation Alignment for Generation: Training Diffusion Transformers Is Easier Than You Think},
  author={Yu, Sihyun and Kwak, Sangkyung and Jang, Huiwon and Jeong, Jongheon and Huang, Jonathan and Shin, Jinwoo and Xie, Saining},
  booktitle={The Thirteenth International Conference on Learning Representations},
  year={2025}
}

@inproceedings{lipmanflow,
  title={Flow Matching for Generative Modeling},
  author={Lipman, Yaron and Chen, Ricky TQ and Ben-Hamu, Heli and Nickel, Maximilian and Le, Matthew},
  booktitle={The Eleventh International Conference on Learning Representations},
  year={2023}
}

@inproceedings{liuflow,
  title={Flow Straight and Fast: Learning to Generate and Transfer Data with Rectified Flow},
  author={Liu, Xingchao and Gong, Chengyue and others},
  booktitle={The Eleventh International Conference on Learning Representations},
  year={2023}
}

@article{raffel2020exploring,
  title={Exploring the limits of transfer learning with a unified text-to-text transformer},
  author={Raffel, Colin and Shazeer, Noam and Roberts, Adam and Lee, Katherine and Narang, Sharan and Matena, Michael and Zhou, Yanqi and Li, Wei and Liu, Peter J},
  journal={Journal of machine learning research},
  volume={21},
  number={140},
  pages={1--67},
  year={2020}
}

@inproceedings{loshchilov2019adamw,
  title     = {Decoupled Weight Decay Regularization},
  author    = {Loshchilov, Ilya and Hutter, Frank},
  booktitle = {International Conference on Learning Representations (ICLR)},
  year      = {2019},
  eprint    = {1711.05101},
  archivePrefix = {arXiv},
  primaryClass = {cs.LG}
}

@inproceedings{karunratanakul2024optimizing,
  title={Optimizing diffusion noise can serve as universal motion priors},
  author={Karunratanakul, Korrawe and Preechakul, Konpat and Aksan, Emre and Beeler, Thabo and Suwajanakorn, Supasorn and Tang, Siyu},
  booktitle={Proceedings of the IEEE/CVF Conference on Computer Vision and Pattern Recognition},
  pages={1334--1345},
  year={2024}
}

@inproceedings{goel2024iterative,
  title={Iterative motion editing with natural language},
  author={Goel, Purvi and Wang, Kuan-Chieh and Liu, C Karen and Fatahalian, Kayvon},
  booktitle={ACM SIGGRAPH 2024 Conference Papers},
  pages={1--9},
  year={2024}
}

@inproceedings{huang2024controllable,
  title={Como: Controllable motion generation through language guided pose code editing},
  author={Huang, Yiming and Wan, Weilin and Yang, Yue and Callison-Burch, Chris and Yatskar, Mark and Liu, Lingjie},
  booktitle={European Conference on Computer Vision},
  pages={180--196},
  year={2024},
  organization={Springer}
}

@inproceedings{raab2024monkey,
  title={Monkey see, monkey do: Harnessing self-attention in motion diffusion for zero-shot motion transfer},
  author={Raab, Sigal and Gat, Inbar and Sala, Nathan and Tevet, Guy and Shalev-Arkushin, Rotem and Fried, Ohad and Bermano, Amit Haim and Cohen-Or, Daniel},
  booktitle={SIGGRAPH Asia 2024 Conference Papers},
  pages={1--13},
  year={2024}
}

@inproceedings{hong2025salad,
  title={Salad: Skeleton-aware latent diffusion for text-driven motion generation and editing},
  author={Hong, Seokhyeon and Kim, Chaelin and Yoon, Serin and Nam, Junghyun and Cha, Sihun and Noh, Junyong},
  booktitle={Proceedings of the Computer Vision and Pattern Recognition Conference},
  pages={7158--7168},
  year={2025}
}

@article{xu2026necromancer,
  title={NECromancer: Breathing Life into Skeletons via BVH Animation},
  author={Xu, Mingxi and Wang, Qi and Wen, Zhengyu and Thien, Phong Dao and Li, Zhengyu and Zhang, Ning and He, Xiaoyu and Zhao, Wei and Gong, Kehong and Zhang, Mingyuan},
  journal={arXiv preprint arXiv:2602.06548},
  year={2026}
}

@article{chen2025topology,
  title={Topology-Agnostic Animal Motion Generation from Text Prompt},
  author={Chen, Keyi and Sun, Mingze and Liu, Zhenyu and Chen, Zhangquan and Huang, Ruqi},
  journal={arXiv preprint arXiv:2512.10352},
  year={2025}
}

@inproceedings{wang2026x,
  title={X-MoGen: Unified Motion Generation Across Humans and Animals},
  author={Wang, Xuan and Ruan, Kai and Qian, Liyang and Zhi, Guo Zhi and Su, Chang and Wang, Gaoang},
  booktitle={Proceedings of the AAAI Conference on Artificial Intelligence},
  volume={40},
  number={12},
  pages={10234--10242},
  year={2026}
}

@article{khani2025unimogen,
  title={Unimogen: Universal motion generation},
  author={Khani, Aliasghar and Rampini, Arianna and Atherton, Evan and Roy, Bruno},
  journal={arXiv preprint arXiv:2505.21837},
  year={2025}
}

@article{ying2021transformers,
  title={Do transformers really perform badly for graph representation?},
  author={Ying, Chengxuan and Cai, Tianle and Luo, Shengjie and Zheng, Shuxin and Ke, Guolin and He, Di and Shen, Yanming and Liu, Tie-Yan},
  journal={Advances in neural information processing systems},
  volume={34},
  pages={28877--28888},
  year={2021}
}

@article{aberman2020skeleton,
  title={Skeleton-aware networks for deep motion retargeting},
  author={Aberman, Kfir and Li, Peizhuo and Lischinski, Dani and Sorkine-Hornung, Olga and Cohen-Or, Daniel and Chen, Baoquan},
  journal={ACM Transactions on Graphics (ToG)},
  volume={39},
  number={4},
  pages={62--1},
  year={2020},
  publisher={ACM New York, NY, USA}
}

@inproceedings{chen2025motion2motion,
  title={Motion2motion: Cross-topology motion transfer with sparse correspondence},
  author={Chen, Ling-Hao and Zhang, Yuhong and Yin, Zixin and Dou, Zhiyang and Chen, Xin and Wang, Jingbo and Komura, Taku and Zhang, Lei},
  booktitle={Proceedings of the SIGGRAPH Asia 2025 Conference Papers},
  pages={1--11},
  year={2025}
}

@article{hu2023motion,
  title={Motion flow matching for human motion synthesis and editing},
  author={Hu, Vincent Tao and Yin, Wenzhe and Ma, Pingchuan and Chen, Yunlu and Fernando, Basura and Asano, Yuki M and Gavves, Efstratios and Mettes, Pascal and Ommer, Bjorn and Snoek, Cees GM},
  journal={arXiv preprint arXiv:2312.08895},
  year={2023}
}

@inproceedings{mo2025pumps,
  title={PUMPS: Skeleton-agnostic point-based universal motion pre-training for synthesis in human motion tasks},
  author={Mo, Clinton Ansun and Hu, Kun and Long, Chengjiang and Yuan, Dong and Siu, Wan-Chi and Wang, Zhiyong},
  booktitle={Proceedings of the IEEE/CVF International Conference on Computer Vision},
  pages={14496--14506},
  year={2025}
}

@article{witkin1988spacetime,
  title={Spacetime constraints},
  author={Witkin, Andrew and Kass, Michael},
  journal={ACM Siggraph Computer Graphics},
  volume={22},
  number={4},
  pages={159--168},
  year={1988},
  publisher={ACM New York, NY, USA}
}

@inproceedings{gleicher1997motion,
  title={Motion editing with spacetime constraints},
  author={Gleicher, Michael},
  booktitle={Proceedings of the 1997 symposium on Interactive 3D graphics},
  pages={139--ff},
  year={1997}
}

@article{sakoe1978dynamic,
  title={Dynamic programming algorithm optimization for spoken word recognition},
  author={Sakoe, Hiroaki and Chiba, Seibi},
  journal={IEEE transactions on acoustics, speech, and signal processing},
  volume={26},
  number={1},
  pages={43--49},
  year={1978},
  publisher={IEEE}
}

@inproceedings{radford2021learning,
  title={Learning transferable visual models from natural language supervision},
  author={Radford, Alec and Kim, Jong Wook and Hallacy, Chris and Ramesh, Aditya and Goh, Gabriel and Agarwal, Sandhini and Sastry, Girish and Askell, Amanda and Mishkin, Pamela and Clark, Jack and others},
  booktitle={International conference on machine learning},
  pages={8748--8763},
  year={2021},
  organization={PmLR}
}

@InProceedings{Shi_2026_CVPR,
    author    = {Shi, Zhenwu and Gong, Jingyu and Wang, Peiwei and Wang, Xingzan and Qian, Tianwen and Li, Wenxi and Fang, Yuan and Xie, Jiao and Ma, Lizhuang and Lin, Shaohui},
    title     = {Omni-Supervised Motion Editing: Balancing Change and Invariance through Positive-Negative Learning},
    booktitle = {Proceedings of the IEEE/CVF Conference on Computer Vision and Pattern Recognition (CVPR)},
    month     = {June},
    year      = {2026},
    pages     = {30595-30606}
}

@InProceedings{Han_2026_CVPR,
    author    = {Han, Gyojin and Kim, Junmo},
    title     = {Cross-Axis Feature Fusion with Joint-Wise Motion Difference Prediction for Text-Based 3D Human Motion Editing},
    booktitle = {Proceedings of the IEEE/CVF Conference on Computer Vision and Pattern Recognition (CVPR)},
    month     = {June},
    year      = {2026},
    pages     = {30618-30628}
}

@InProceedings{Petrovich_2023_ICCV,
    author    = {Petrovich, Mathis and Black, Michael J. and Varol, G\"ul},
    title     = {TMR: Text-to-Motion Retrieval Using Contrastive 3D Human Motion Synthesis},
    booktitle = {Proceedings of the IEEE/CVF International Conference on Computer Vision (ICCV)},
    month     = {October},
    year      = {2023},
    pages     = {9488-9497}
}

@inproceedings{singh2025matters,
  title={What matters for representation alignment: Global information or spatial structure?},
  author={Singh, Jaskirat and Leng, Xingjian and Wu, Zongze and Zheng, Liang and Zhang, Richard and Shechtman, Eli and Xie, Saining},
  booktitle={The Fourteenth International Conference on Learning Representations},
  year={2025}
}

@inproceedings{zheng2026diffusion,
  title={Diffusion transformers with representation autoencoders},
  author={Zheng, Boyang and Ma, Nanye and Tong, Shengbang and Xie, Saining},
  booktitle={International Conference on Learning Representations},
  volume={2026},
  pages={35791--35820},
  year={2026}
}

@inproceedings{shin2026representation,
  title={Representation alignment for just image transformers is not easier than you think},
  author={Shin, Jaeyo and Kim, Jiwook and Shim, Hyunjung},
  booktitle={European Conference on Computer Vision},
  pages={603--620},
  year={2026},
  organization={Springer}
}

@article{wang2026persistence,
  title={Persistence Forcing: Exploiting Feature Specialization in Pixel-Space Diffusion},
  author={Wang, Chong and Fu, Zixuan and Huang, Shiqi and Yang, Siyuan and Cheng, Hao and Wen, Bihan},
  journal={arXiv preprint arXiv:2609.36014},
  year={2026}
}
